\documentclass[11pt,a4paper]{article}
\PassOptionsToPackage{breaklinks}{hyperref}
\usepackage[hyperref]{acl2019}
\usepackage{times}
\usepackage{tabularx}
\usepackage{latexsym}
\usepackage{multirow}
\usepackage{breakurl}
\usepackage{graphicx}
\usepackage{algorithm2e}

\aclfinalcopy 

\title{CUNI System for the Building Educational Applications 2019\\Shared Task: Grammatical Error Correction}

\author{Jakub N\'{a}plava \and Milan Straka\\
  Charles University, \\
  Faculty of Mathematics and Physics, \\
  Institute of Formal and Applied Linguistics \\
  \texttt{\{naplava,straka\}@ufal.mff.cuni.cz}}

\date{}

\begin{document}
\maketitle
\begin{abstract}
  In this paper, we describe our systems submitted to the Building Educational Applications (BEA) 2019 Shared Task~\citep{bea2019}. We participated in all three tracks. Our models are NMT systems based on the Transformer model, which we improve by incorporating several enhancements: applying dropout to whole source and target words, weighting target subwords, averaging model checkpoints, and using the trained model iteratively for correcting the intermediate translations. The system in the Restricted Track is trained on the provided corpora with oversampled ``cleaner'' sentences and reaches 59.39 F0.5 score on the test set. The system in the Low-Resource Track is trained from Wikipedia revision histories and reaches 44.13 F0.5 score. Finally, we finetune the system from the Low-Resource Track on restricted data and achieve 64.55 F0.5 score, placing third in the Unrestricted Track.

\end{abstract}

\section{Introduction}

Starting with the 2013 and 2014 CoNLL Shared Tasks on grammatical error correction (GEC), much progress has been done in this area. The need to correct a variety of error types lead most researchers to focus on models based on machine translation \cite{brockett2006correcting} rather than custom designed rule-based models or a combination of single error classifiers. The machine translation systems turned out to be particularity effective when~\newcite{junczys2016phrase} presented state-of-the-art statistical machine translation system. Currently, models based on statistical and neural machine translation achieve best results: in restricted settings with training limited to certain public training sets \cite{zhao2019improving}; unrestricted settings with no restrictions on training data \cite{ge2018reaching}; and also in low-resource track where the training data should not come from any annotated corpora \cite{lichtarge2018weakly}.\footnote{Note that in this settings Wikipedia revisions are allowed}

In this paper, we present our models and their results in the restricted, unrestricted, and low-resource tracks. We start with a description of related work in Section~\ref{sec:related_work}. We then describe our systems together with the implementation details in Section~\ref{sec:our_system}. Section~\ref{sec:results} is dedicated to our results and  ablation experiments. Finally, in Section~\ref{sec:conclusion} we conclude the paper with some proposals on future work.

\section{Related Work}
\label{sec:related_work}

Transformer \cite{vaswani2017attention} is currently one of the most popular architectures used in machine translation. Its self-attentive layers allow better gradient flow when compared to recurrent neural models and the masking in decoder provides faster training. \newcite{junczys2018approaching} propose several improvements for training Transformer on GEC: using dropout on whole input words, assigning weight to target words based on their alignment to source words, and they also propose to oversample sentences from the training set in order to have the same error rate as the~test~set. 

Majority of work in grammatical error correction has been done in restricted area with a fixed set of annotated training datasets. \newcite{lichtarge2018weakly}, however, show that training a neural machine translation system from Wikipedia edits can lead to surprisingly good results. As the authors state, corpus of Wikipedia edits is only weakly supervised for the task of GEC, because most of the edits are not corrections of grammatical errors and also they are not human curated specifically for GEC. To overcome these issues, the authors use iterative decoding which allows for incremental corrections. In other words, the model can repeatedly translate its current output as long as the translation is more probable then keeping the sentence unchanged. Similar idea is also presented in \cite{ge2018reaching}, where the translation system is trained with respect to the incremental inference.

\section{Our System}
\label{sec:our_system}

In this section, we present our three systems submitted to each track of the BEA 2019 Shared Task. We start with the Restricted Track In Section~\ref{ssec:restricted_track}, where we present a series of improvements to the baseline Transformer model. In Section~\ref{ssec:low_resource_track}, we describe our model trained on Wikipedia revisions which was submitted to the Low-Resource Track. Finally, in Section~\ref{ssec:unrestricted_track}, we describe the model submitted to the Unrestricted Track.

All our models are based on the Transformer model from Tensor2Tensor framework version~1.12.0.\footnote{https://github.com/tensorflow/tensor2tensor}

\subsection{Restricted Track}
\label{ssec:restricted_track}

In the Restricted Track, we use the 5 provided datasets for system development: FCE v2.1 \citep{yannakoudakis2011new}, Lang-8 Corpus of Learner English \citep{mizumoto2011mining, tajiri2012tense}, NUCLE \citep{dahlmeier2013building}, Write \& Improve (W\&I) and LOCNESS v2.1 \citep{bea2019, granger1998}. From Lang-8 corpus, we took only the sentences annotated by annotators with ID 0 (A0) and ID 1 (A1). All but the development sets from W\&I and LOCNESS datasets were used for training. The simple statistics of these datasets are presented in Table~\ref{table:restricted_data}. The displayed error rate is computed using maximum alignment of original and annotated sentences as a ratio of non-matching alignment edges (insertion, deletion, and replacement).

We use the \textit{transformer\_base} configuration of Tensor2Tensor as our baseline solution. The training dataset consists of 1\,230\,231 sentences. After training, beam search decoding is employed to generate model corrections and we choose the checkpoint with the highest accuracy on a development set concatenated from the W\&I and LOCNESS development sets.

\begin{table}[t]
  \begin{center}
    \begin{tabular}{l|c||r|r}
      \multicolumn{2}{c||}{\multirow{2}{*}{Dataset}} & \multirow{2}{*}{Sentences} & Average \\
      \multicolumn{2}{c||}{}                         &                            & error rate \\\hline\hline
      \multirow{2}{*}{Lang8}
        & A0  & 1\,037\,561 & 13.33 \% \\\cline{2-4}
        & A1 & 67\,975 & 25.84 \%\\\hline
      \multirow{3}{*}{FCE v2.1}
        & train  & 28\,350 & 11.31 \%  \\\cline{2-4}
        & dev & 2\,191 & 11.67 \%  \\\cline{2-4}
        & test & 2\,695 & 12.87 \%  \\\hline
      NUCLE
        &   & 57\,151 & 6.56 \%  \\\hline
      \multirow{6}{*}{W\&I}
        & train A & 10\,493 & 18.13 \%  \\\cline{2-4}
        & train B & 13\,032 & 11.68 \%  \\\cline{2-4}
        & train C & 10\,783 & 5.62 \%   \\\cline{2-4}
        & dev A & 1\,037 & 18.32 \%  \\\cline{2-4}
        & dev B & 1\,290 & 12.46 \%  \\\cline{2-4}
        & dev C & 1\,069 & 5.91 \%  \\\hline
      LOCNESS
        & dev N & 998 & 4.72 \%  \\\hline
    \end{tabular}
    \caption{Statistics of available datasets. The error rate is computed as a ratio of non-matching alignment edges.}
    \label{table:restricted_data}
  \end{center}
\end{table}

\subsubsection{Transformer Big}

The first minor improvement was to use the \textit{transformer\_big} configuration instead of \textit{transformer\_base}. This configuration has bigger capacity and as \newcite{popel2018training} show, it reaches substantially better results on certain translation tasks.

\subsubsection{Source and Target Word Dropout}

Dropout~\cite{srivastava2014dropout} is a regularization technique that turned out to be particularly effective in the field of neural networks. It works by masking several randomly selected activations during training, which should prevent the neural network from overfitting the training data. In the area of NLP, it is a common approach to apply dropout to whole embeddings, randomly zeroing certain dimensions. As \newcite{junczys2018approaching} show, we can also apply dropout to whole source words to reduce trust in the source words. Specifically, full source word embedding vector is set to zero vector with probability $p$. We further note this probability as the \textit{source\_word\_dropout}.

To make regularization even more effective, we decided to dropout also whole target word embeddings. We refer to the probability with which we dropout entire target word embeddings as the \textit{target\_word\_dropout}.

\subsubsection{Edited MLE}

Compared to traditional machine translation task, whose goal is to translate one language to another, GEC operates on a single language. Together with the relatively low error rate, the translation system may converge to a local optimum, in which the model copies the input unchanged to the output. To overcome this issue, \newcite{junczys2018approaching} propose to change the maximum likelihood objective to assign bigger weights to target tokens different from the source tokens. More specifically, they start by computing the word alignment between each source $x=(x_0, x_1,..x_N)$ and target sentence $y=(y_0, y_1,...y_M)$. Then they set the weight $\lambda_t$ of the target word $y_t$ to 1 if it is matched, and otherwise, if it is an insertion or replacement of a source token, $\lambda_t$ is set to some predefined constant. Modified log-likelihood training objective then takes following form:
$$ L(x,y) = - \sum_{t=1}^{M} \lambda_t \log P(y_t|x,y_0,\ldots, y_{t-1}). $$

\subsubsection{Data oversampling}
\label{ssec:data_oversample}

It is crucial to have training data from the same domain as the test data, i.e., training data containing similar errors with similar distribution as the test data. As we can see in the Table~\ref{table:restricted_data}, the vast majority of our training data comes from the Lang-8 corpus. However, as it is quite noisy and of low quality, it matches the target domain the least. Therefore, we decided to oversample other datasets. Specifically, we add the W\&I training data 10 times, all FCE data 5 times and NUCLE corpus 5 times to the training data. The oversampled training set consists of 1\,900\,551.

In Table~\ref{table:restricted_data}, we can also see token error rate of each corpus. The development error rate in W\&I and LOCNESS varies from 5.91\% up to 18.32\%. This gives us a basic idea how the test data looks like, and since the test data does not contain annotations from which set (\verb|A|, \verb|B|, \verb|C|, \verb|N|) it comes, we decided not to optimize the training data against the token error rate any further.

\subsubsection{Checkpoint Averaging}

\newcite{popel2018training} report that averaging several last Transformer model checkpoints during training leads both to lower variance results and also to slightly better performance than the baseline without averaging. They propose to save checkpoints every one hour and average either 8 or 16 last checkpoints. Since we found out that the model overfits the oversampled dataset quite quickly, we save checkpoints every 30 minutes.

\subsubsection{Iterative decoding}

\begin{algorithm}[]
 \SetAlgoLined
 \KwData{$\textit{input\_sent};\ \textit{max\_iters};\ \textit{threshold}$ }
 \For{$\textit{iter}$ in [1,2,..,\textit{max\_iters}]} {
        $\textit{beam\_results}$ = decode($\textit{input\_sent}$)\;

        $\textit{identity\_cost}$ =  $+ \infty$\;
        $\textit{non\_identity\_cost}$ = $+ \infty$\;
        $\textit{non\_identity\_sent}$ = None

        \For{$\textit{beam\_item}$ in $\textit{beam\_results}$} {
                $\textit{text}$ = $\textit{beam\_item}[\textrm{"text"}]$\;
                $\textit{cost}$ = $\textit{beam\_item}[\textrm{"cost"}]$\;
                \uIf{$\textit{text}$ == $\textit{input\_sent}$}{
                        $\textit{identity\_cost}$ = $\textit{cost}$\;
                }
                \uElseIf{ $\textit{cost} < \textit{non\_identity\_cost}$} {
                        $\textit{non\_identity\_cost}\ =\ \textit{cost}$\;
                        $\textit{non\_identity\_sent}\ = \ \textit{text}$\;
                }
        }

        \eIf{$\textit{non\_identity\_cost} \le \textit{threshold} \cdot \textit{identity\_cost}$}{
                $\textit{input\_sent}$ = $\textit{non\_identity\_sent}$\;
        } {
                break\;
        }
 }
 return $\textit{input\_sent}$\;

 \caption{Iterative decoding algorithm}
 \label{alg:iterative_decoding}
\end{algorithm}

A system for grammatical error correction should correct all errors in the text while keeping the rest of the text intact. In many situations with multiple errors in a sentence, the trained system, however, corrects only a subset of its errors. \newcite{lichtarge2018weakly} and \newcite{ge2018reaching} propose to use the trained system iteratively to allow the system to correct certain errors during further iterations. Iterative decoding is done as long as the cost of the correction is less than the cost of the identity translation times a predefined constant. While \newcite{lichtarge2018weakly} use the same trained model log-likelihoods as the cost function, \newcite{ge2018reaching} utilize an external language model for it. Because the restricted track does not contain enough training data to train a quality language model, we adopted the first approach and utilize the trained system log-likelihoods as a stopping criterion.

The iterative decoding algorithm we use is presented in Algorithm~\ref{alg:iterative_decoding}. Note that when the resulting beam does not contain the identical (non-modified) sentence, the correction with the lowest cost is returned regardless of the provided threshold. We adopted this approach for two reasons -- efficiently obtaining the log-likelihood of the identical sentence would require non-trivial modification of the Tensor2Tensor framework, and for $\textit{threshold} > 1$ (i.e., allow generating changes which are less likely than identical sentence) the results are the same.


\subsubsection{Implementation Details}
\label{ssec:restricted_track_implementation_details}

Apart from the first experiment in which we use \textit{transformer\_base} configuration, all our experiments are based on \textit{transformer\_big} architecture. We use Adafactor optimizer~\cite{shazeer2018adafactor}, linearly increasing the learning rate from 0 to 0.011 over the first 8000 steps, then decrease it proportionally to the number of steps after that.\footnote{We use 8000 warmup steps and learning\_rate\_schedule=rsqrt\_decay} We also experimented with Adam optimizer with default learning rate schedule, however, training converged poorly. We hypothesise that this was caused by the higher learning rate.

All systems are trained on 4 Nvidia P5000 GPUs for approximately 2 days. The vocabulary consists of approximately 32k most common word-pieces, batch size is 2000 word-pieces per each GPU and all sentences with more than 150 word-pieces are discarded. Model checkpoints are saved every 30 minutes. We ran a grid search to find values of all hyperparameters described in the previous sections.

At evaluation time, we run iterative decoding using a beam size of 4. Beam-search length-balance decoding hyperparameter alpha is set to 0.6. This applies to all further experiments.

\subsection{Low-Resource Track}
\label{ssec:low_resource_track}

The dataset for our experiments in the Low-Resource Track consists of nearly 190M segment pairs extracted from Wikipedia XML revision dumps. To acquire these, we downloaded all English Wikipedia revision dumps (155GB in size) and processed them with the \textit{WikiRevision} dataset problem from Tensor2Tensor. The processing pipeline extracts individual pages with chronological snapshots, removes all non-text elements and downsamples the snapshots. With low probability, additional spelling noise is added by either inserting a random character, deleting a random character, transposing two adjacent characters or replacing a character with a random one. With the same low probability, a random text substring (up to 8 characters) may also be replaced with a marker, which should force the model to learn infilling. Finally, the texts from two consecutive snapshots are aligned and sequences between matching segments are extracted to form a training pair. Only 4\% of identical samples are preserved.

Despite having an enormous size compared to 1.2M sentences in the Restricted Track, the training pairs extracted from Wikipedia are extremely noisy, containing a lot of edits that are in no sense grammatical correction. It is also worth noting that the identical data modified by the spelling and infilling operations form nearly 50\% of the training pairs.

Since we want to re-use the system in other scenarios, we train the model on the original (untokenized) training data. To evaluate the model on the BEA development and test data, we detokenize the data using Moses,\footnote{We use mosestokenizer v1.0.0 and its detokenizer.} run model inference and finally tokenize corrected sentences using spaCy.\footnote{We use spaCy v1.9.0 and the en\_core\_web\_sm-1.2.0 model.}

The training segments may contain newline and tab symbols; therefore, we applied additional post-processing in which we replaced both these symbols with spaces.

Because overfitting should not be an issue with the Wikipedia data, we decided to use \textit{transformer\_clean\_big\_tpu} configuration, following \newcite{lichtarge2018weakly}. This configuration, compared to \textit{transformer\_big}, performs no dropouts. The vocabulary consists of approximately 32k most common word-pieces, batch size is 2000 word-pieces per each GPU and all sentences with more than 150 word-pieces are discarded. We train the model for approximately 10 days on 4 Nvidia P5000 GPUs. After training, the last 8 checkpoints saved in 1 hour intervals are averaged. Finally, we run a grid search to find optimal values of \textit{threshold} and \textit{max\_iters} in iterative decoding algorithm.

\begin{table*}[t]
  \begin{center}
    \begin{tabular}{l|c|c|c|c|r}
      Track & P & R & F$_{0.5}$ & Best & Rank \\\hline\hline
      Restricted & 67.33 & 40.37 & 59.39 & 69.47 & 10 / 21 \\\hline
      Unrestricted & 68.17 & 53.25 & 64.55 & 66.78 &  3 / \hphantom{0}7 \\\hline
      Low Resource & 50.47 & 29.38 & 44.13 & 64.24 &  5 / \hphantom{0}9 \\\hline
    \end{tabular}
  \end{center}
  \caption{Official shared task F$_{0.5}$ scores on the test set.}
  \label{table:st_official_results}
\end{table*}

\begin{table*}
  \begin{center}
    \begin{tabular}{l|c|c|c|c|c}
      System & A     & B     & C     & N     & Combined \\ \hline\hline
      Transformer-base architecture       & 39.98 & 32.68 & 23.97 & 14.49 & 32.47    \\ \hline
      Transformer-big architecture        & 39.70 & 35.13 & 26.22 & 20.20 & 34.20    \\ \hline
      + 0.2 src drop, 0.1 tgt drop, 3 MLE & 42.06 & 38.25 & 28.72 & 23.80 & 38.15    \\ \hline
      ~~~ + Extended dataset              & 45.99 & 41.79 & 32.52 & 27.89 & 40.86    \\ \hline
      ~~~~~~~ + Averaging 8 checkpoints    & 47.90  & 44.13 & 36.19 & 29.05 & 43.29    \\ \hline
      ~~~~~~~~~~~ + Iterative decoding    & 48.75 & 45.46 & 37.09 & 30.19 & 44.27    \\ \hline
    \end{tabular}
  \end{center}
  \caption{Development combined F$_{0.5}$ score of incremental improvements of our system.}
  \label{table:restricted_results}
\end{table*}

\subsection{Unrestricted Track}
\label{ssec:unrestricted_track}

Our system submitted to the Unrestricted Track is the best system from the Low-Resource Track finetuned on the oversampled training data as described in Section~\ref{ssec:data_oversample}. Since our system in the Unrestricted Track was trained on detokenized data, the training sentences for finetuning were also detokenized. The tokenization and detokenization was done in the same way as described in Section~\ref{ssec:low_resource_track}.

We finetune the system with the Adafactor optimizer. The learning rate linearly increases from 0 to 0.0003 over the first 20\,000 steps and then remains constant. We employ source word dropout, target word dropout and weighted MLE. The training data for finetuning and the rest of the training scheme are identical to Section~\ref{ssec:restricted_track_implementation_details}.

\section{Results}
\label{sec:results}

We now present the results of our system. Additionally, we present several
ablation experiments, which are evaluated on the concatenation of W\&I and LOCNESS
development sets (the \emph{Dev combined}).


\subsection{Shared Task Results}

The official results of our three systems on the blind test set are presented in Table~\ref{table:st_official_results}. All our systems have substantially higher precision than recall. It is an interesting observation that the system in the unrestricted track has similar precision as the model in the restricted track while having higher recall.

\subsection{Restricted Track}

The first experiment we conducted is devoted to the incremental enhancements that we proposed in Section~\ref{ssec:restricted_track}. As Table~\ref{table:restricted_results} indicates, applying each enhancement results in higher performance on the development set. By applying all incremental improvements, total $F_{0.5}$ score on the development set increases by 11.8\%.

We improved the $F_{0.5}$ score by adding \textit{source\_word\_dropout}, \textit{target\_word\_dropout} and MLE weighting by almost 4\%. To find out optimal values of all three hyper-parameters, we ran a small grid search. The results of this experiment are presented in Table~\ref{table:restricted_ablations}. The source-word dropout improves the results the most, MLE provides minor gains, while the influence of target-word dropout on the results is unclear.

\begin{table}
  \begin{center}
    \begin{tabular}{l|l|l||c}
      \multicolumn{1}{c|}{Source}  & \multicolumn{1}{c|}{Target}  & \multirow{3}{*}{MLE} & Dev       \\
      \multicolumn{1}{c|}{word}    & \multicolumn{1}{c|}{word}    &                      & combined  \\
      \multicolumn{1}{c|}{dropout} & \multicolumn{1}{c|}{dropout} &                      & F$_{0.5}$ \\\hline\hline
      0   & 0   & 1  & 34.20 \\ \hline
      0.1 &     &    & 37.89 \\ 
      0.2 &     &    & 38.26 \\ \hline
          & 0.1 &    & 35.43 \\
          & 0.2 &    & 33.98 \\ \hline
          &     & 2  & 34.56 \\
          &     & 3  & 34.28 \\
          &     & 4  & 34.17 \\ \hline
      0.2 & 0.1 &    & 37.89 \\ \hline
      0.2 &     & 3  & 38.68 \\ \hline
      0.2 & 0.1 & 3  & 38.15 \\ \hline
    \end{tabular}
  \end{center}
  \caption{The effect of source word dropout, target word dropout, and MLE
  weight on development combined F$_{0.5}$ score.}
  \label{table:restricted_ablations}
\end{table}

\begin{figure}[t]
  \begin{center}
    \includegraphics[width=\hsize]{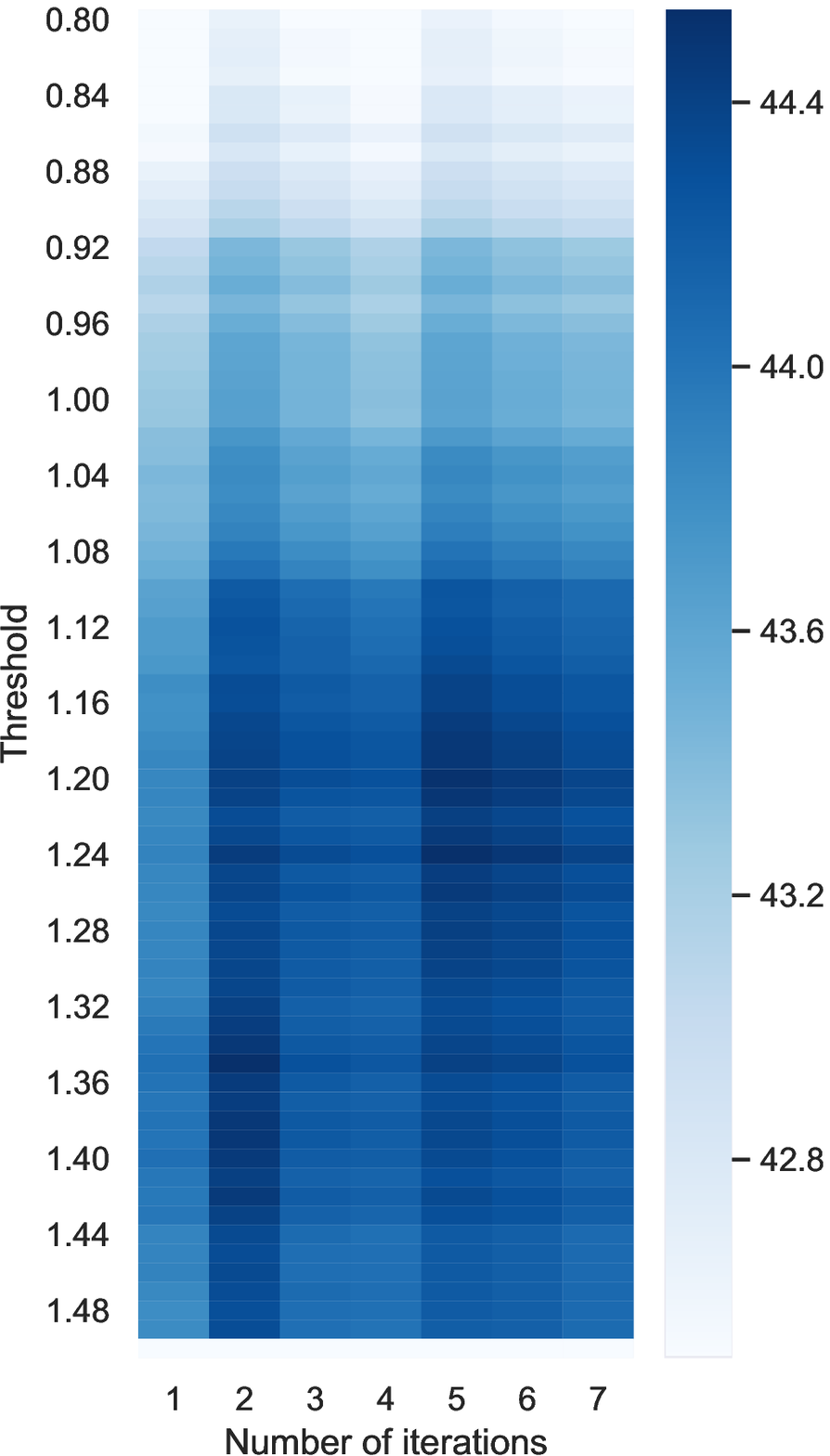}
  \end{center}
  \caption{Performance of iterative decoding depending on number of iterations
    and threshold parameters.}
  \label{fig:restriced_iterative_decoding}
\end{figure}

\begin{table}
  \begin{center}
    \begin{tabular}{l||c}
      Checkpointing & Dev combined F$_{0.5}$ \\\hline\hline
      No checkpointing & 41.55 \\\hline
      Averaging 4 checkpoints & 43.00 \\\hline
      Averaging 6 checkpoints & 43.13 \\\hline
      Averaging 8 checkpoints & 43.29 \\\hline
    \end{tabular}
  \end{center}
  \caption{Maximum development combined F$_{0.5}$ score achieved by averaging the given number of checkpoints.}
  \label{table:restricted_checkpoint_averaging}
\end{table}

In the next experiment, we examined the effect of checkpoint averaging. Table~\ref{table:restricted_checkpoint_averaging} presents results of the model without averaging and with averaging 4, 6, and 8 model checkpoints. The best results are achieved when 8 checkpoints are used and the results indicate that the more checkpoints are averaged the better the results are.

Finally, we inspect the effect of iterative decoding. Specifically, we run an exhaustive grid search to find optimal values of \textit{threshold} and \textit{max\_iters}. The results of this experimented are visualised in Figure~\ref{fig:restriced_iterative_decoding}. We can see that increasing \textit{threshold} from 1 to values around 1.20 leads to substantially better results. Moreover, using more iterations also has a positive impact on the model performance. Both of these improvements are caused by the model generating more corrections which are deemed less likely to the model, i.e., we increase recall at the expense of precision.

\subsection{Low-Resource Track}

We train following models in the Low-Resource Track:
\begin{enumerate}

\item the \textit{transformer\_big} configuration with \textit{input\_word\_dropout} set to 0.2 and \textit{target\_word\_dropout} to 0.1 -- settings similar to the best system in the Restricted Track but without edited MLE;
\item the \textit{transformer\_clean\_big\_tpu} configuration -- this configuration uses no internal dropouts;
\item the \textit{transformer\_clean\_big\_tpu} configuration with \textit{input\_word\_dropout} 0.2 and \textit{target\_word\_dropout} 0.1;
\item the \textit{transformer\_clean\_big\_tpu} configuration trained on sentences extracted from Wikipedia revisions without introducing additional spelling errors and infillment marker.
\end{enumerate}

\begin{table}
  \begin{center}
    \setlength{\tabcolsep}{4pt}
    \begin{tabular}{l|l||c}
      \multirow{3}{*}{ID} & \multirow{3}{*}{Model} & Dev \\
      & & combined \\
      & & F$_{0.5}$ \\\hline\hline
      \multirow{2}{*}{1} & \textit{transformer\_big} & \multirow{2}{*}{22.03} \\
                         & 0.2 src drop, 0.1 tgt drop & \\\hline
      \multirow{2}{*}{2} & \textit{transformer\_clean\_big\_tpu} & \multirow{2}{*}{26.05} \\
                         & no src drop, no tgt drop & \\\hline
      \multirow{2}{*}{3} & \textit{transformer\_clean\_big\_tpu} & \multirow{2}{*}{24.80} \\
                         & 0.2 src drop, 0.1 tgt drop & \\\hline
      \multirow{2}{*}{4} & \textit{transformer\_clean\_big\_tpu} & \multirow{2}{*}{21.16} \\
                         & no spelling or infillment errors & \\\hline
    \end{tabular}
  \end{center}
  \caption{Development combined F$_{0.5}$ score achieved with different models in the Low-Resource Track.}
  \label{table:low_resource_track_ablations}
\end{table}

All but the fourth model use the training data as described in Section~\ref{ssec:low_resource_track} and the training scheme is in all models identical. The results of all models are presented in Table~\ref{table:low_resource_track_ablations}.

The best results are achieved with the second model which performs no dropouts. When we incorporate source and target word dropouts in the third experiment, the performance deteriorates by more than 1\%. When we also add Transformer internal dropouts in the first experiment, the performance drops by additional 2.8\%. This confirms our assumption that the enormous amount of data is strong enough regularizer and the usage of additional regularizers leads to worse performance.

The results of the fourth model, which was trained on data without additional spelling and infillment noise, are almost 5\% worse than when training on data with this noise. It would be an interesting experiment to evaluate the effect of spelling and infillment noise separately, but this was not done in this paper.

\begin{figure}
  \begin{center}
    \includegraphics[width=\hsize]{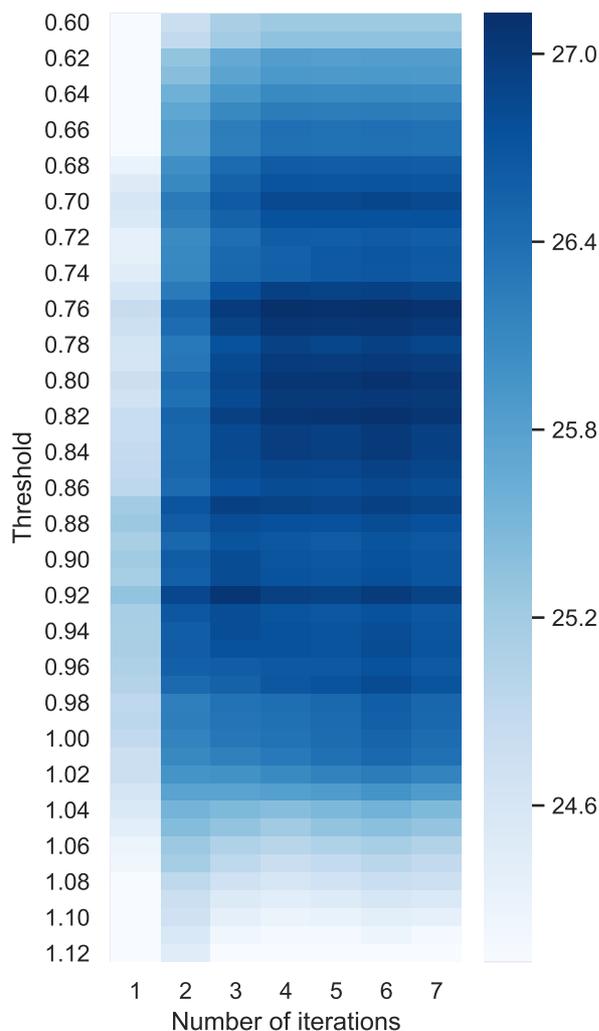}
  \end{center}
  \caption{Performance of iterative decoding depending on number of iterations
    and threshold parameters.}
  \label{fig:wiki_iterative_decoding}
\end{figure}

We also run an exhaustive grid search to find optimal values of \textit{threshold} and \textit{max\_iters} in iterative decoding. As we can see in Figure~\ref{fig:wiki_iterative_decoding}, the optimal value of \textit{threshold} is now below 1 indicating that precision is now increased at the expense of recall. A performance gain in using more than one iteration is clearly visible.

\subsection{Unrestricted Track}

In the Unrestricted Track, we tried finetuning the pretrained system with two different learning rate schedules:

\begin{itemize}
\item linearly increase learning rate from 0 to 0.011 over the first 8000 steps, then decrease it proportionally to the number of steps after that -- exactly same as while training system from scratch in the Restricted Track (see Section~\ref{ssec:restricted_track_implementation_details});
\item linearly increase learning rate from 0 to 3e-4 then keep the learning rate constant as proposed by \newcite{lichtarge2018weakly}.
\end{itemize}

All other hyper-parameters and the training process remain the same as described in Section~\ref{ssec:unrestricted_track}.

The first finetuning scheme overfitted the training corpus quite quickly while reaching score of 48.33. The second scheme converged slower and reached a higher score of 48.82.

\section{Conclusion}
\label{sec:conclusion}

We have presented our three systems submitted to the BEA 2019 Shared Tasks. By
employing larger architecture, source and target word dropout, edited MLE,
dataset extension, checkpoint averaging, and iterative decoding, our system
reached 59.39 F$_{0.5}$ score in the Restricted Track, finishing
10$^\mathrm{th}$ out of 21 participants.

In the Low Resource Track, we utilized Wikipedia revision edits as a training
data, reaching 44.14 F$_{0.5}$ score. Finally, we finetuned this model using
the annotated training data, obtaining 65.55 F$_{0.5}$ score in the Unrestricted Track,
ranking 3$^\mathrm{rd}$ out of 7 submissions.

As future work, we would like to explore iterative decoding algorithm more thoroughly. Specifically, we hope that allowing \textit{threshold} parameter to change in each iteration might provide gains. We would also like to train systems on Wikipedia revisions in other languages.

\section*{Acknowledgements}

The work described herein has been supported by OP VVV VI LINDAT/CLARIN project
(CZ.02.1.01/0.0/0.0/16\_013/0001781) and it has been supported and has been
using language resources developed by the LINDAT/CLARIN project (LM2015071) of
the Ministry of Education, Youth and Sports of the Czech Republic.
This research was also partially supported by SVV project number 260~453
and GAUK 578218 of the Charles University.



\bibliography{acl2019}
\bibliographystyle{acl_natbib}

\end{document}